\pdfoutput=1
\documentclass[11pt]{article}

\usepackage{ACL2023}

\usepackage{times}
\usepackage{latexsym}
\usepackage{url}
\usepackage{enumitem}
\usepackage{tcolorbox}
\usepackage{colortbl}
\usepackage{diagbox}
\usepackage{geometry}
\usepackage{booktabs}
\usepackage{multirow}
\usepackage[ruled,boxed,vlined]{algorithm2e}
\usepackage{caption}
\usepackage{graphicx}
\usepackage{subfig}
\usepackage{float} 
\usepackage{subcaption}
\usepackage{braket}
\usepackage{amsmath}
\usepackage{float}
\newcommand{\partitle}[1]{\smallskip \noindent \textbf{#1.}}
\usepackage{overpic}
\usepackage{stfloats}
\usepackage{graphicx}
\usepackage{amsmath}
\usepackage{amsfonts}
\usepackage{hyperref}
\usepackage[T1]{fontenc}
\usepackage[utf8]{inputenc}
\usepackage{microtype}
\usepackage{inconsolata}
\title{\textit{LinkPrompt}: Natural and Universal Adversarial Attacks on Prompt-based Language Models}
\author{Yue Xu \and Wenjie Wang  \Thanks{ W.Wang is the corresponding author.} \\
          School of Information Science and Technology, ShanghaiTech University\\
         \texttt{\{xuyue2022,wangwj1\}@shanghaitech.edu.cn}}
\begin{document}
\maketitle
\begin{abstract}
Prompt-based learning is a new language model training paradigm that adapts the Pre-trained Language Models (PLMs) to downstream tasks, which revitalizes the performance benchmarks across various natural language processing (NLP) tasks. Instead of using a fixed prompt template to fine-tune the model, some research demonstrates the effectiveness of searching for the prompt via optimization. Such prompt optimization process of prompt-based learning on PLMs also gives insight into generating adversarial prompts to mislead the model, raising concerns about the adversarial vulnerability of this paradigm. Recent studies have shown that universal adversarial triggers (UATs) can be generated to alter not only the predictions of the target PLMs but also the prediction of corresponding Prompt-based Fine-tuning Models (PFMs) under the prompt-based learning paradigm. However, UATs found in previous works are often unreadable tokens or characters and can be easily distinguished from natural texts with adaptive defenses. In this work, we consider the naturalness of the UATs and develop \textit{LinkPrompt}, an adversarial attack algorithm to generate UATs by a gradient-based beam search algorithm that not only effectively attacks the target PLMs and PFMs but also maintains the naturalness among the trigger tokens. Extensive results demonstrate the effectiveness of \textit{LinkPrompt}, as well as the transferability of UATs generated by \textit{LinkPrompt} to open-sourced Large Language Model (LLM) Llama2 and API-accessed LLM GPT-3.5-turbo. The resource is available at \href{https://github.com/SavannahXu79/LinkPrompt}{https://github.com/SavannahXu79/LinkPrompt}.

\end{abstract}
\vspace{-1em}
\section{Introduction}
\begin{figure}[t]
\setlength{\abovecaptionskip}{0.2cm}
  \centering
  \includegraphics[width=.9\linewidth]{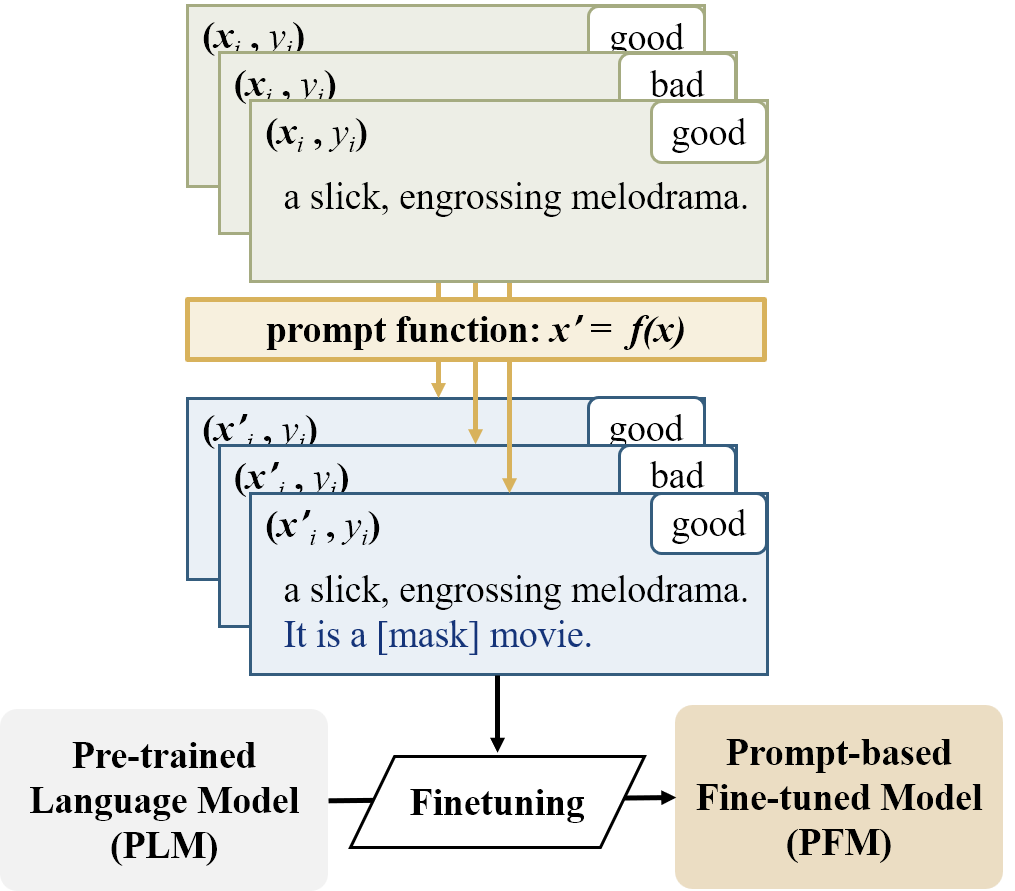}
  \caption{The illustration of prompt-based learning. }
  \label{fig: prompt}
\vspace{-1.5em}
\end{figure}

\vspace{-0.5em}
%\footnote{\url{https://2023.aclweb.org/calls/main_conference/}} 
%A new paradigm called “prompt-based learning” has recently refreshed the state-of-the-art performance in diverse natural language processing tasks \cite{petroni2019language, radford2019language, brown2020language, schick2020s}. 
Prompt-based learning is a new language model training paradigm that aims to adapt the Pre-trained Language Models (PLMs) to perform well on the downstream tasks, which revitalizes the performance benchmarks across various natural language processing (NLP) tasks \cite{radford2019language, brown2020language, schick2020s}. By equipping input sentences with designed prompt templates \cite{liu2023pre}, prompt-based learning converts a text classification task into a next-word prediction task. Then the PLMs are fine-tuned under the prompt-based learning framework to get Prompt-based Fine-tuned Models (PFMs) that are specific to downstream tasks. Specialized prompts can effectively connect PLMs with downstream tasks in few-shot scenarios \cite{winata2021language, tsimpoukelli2021multimodal}. The process of prompt-based learning is demonstrated in Figure \ref{fig: prompt}.

To further enhance the performance of PLMs and PFMs, instead of using a fixed prompt template to fine-tune the model, some methods are proposed to optimize the prompts by maximizing the probability of desired outcomes. 
For example, AutoPrompt \cite{shin2020autoprompt} applied a gradient-based search strategy to optimize a universal prompt template with a fixed length of tokens specific to a downstream task, thus improving the model training efficiency and the generalization ability.

However, such a prompt optimization process of prompt-based learning on PLMs also gives insight into generating adversarial prompts that can mislead the model predictions. Adversarial examples were first discovered and studied in the image domain, that a well-trained image classification model can be easily fooled by adding unnoticeable perturbation to the input space \cite{szegedy2013intriguing, goodfellow2014explaining}. Further studies have shown that such adversarial examples also exist in the text domain, which can be designed by manipulating the words or characters under certain semantic and syntactic constraints \cite{ren2020generating, jin2019bert, zang2020word}. 

Similar to the adversarial attack on simple text classification models, PLMs under prompt-based learning frameworks also suffer from potential adversarial threats. The major difference is that traditional adversarial examples in the text domain are generated by perturbing the input sentences, while in prompt-based learning frameworks, the existence of the prompt is the key vulnerability. \citet{wallace2019universal} first propose a universal adversarial attack on PLMs by optimizing universal adversarial triggers (UATs) that can cause a model to give wrong predictions to any inputs. 

Furthermore, recent studies discovered that the vulnerability to adversarial attacks of PLMs can also be carried to the PFMs under the prompt-based learning paradigm.
\citet{xu2022exploring} proposed a universal adversarial attack named AToP under the prompt-based learning paradigm and proved that the adversarial trigger optimized to target the PLMs can also transfer to the PFMs. Although AToP can successfully diminish the prediction accuracy of PFMs, such UATs have a limitation in naturalness, which means they are meaningless combinations of tokens and symbols that adaptive defense techniques with simple heuristics can easily detect. 

The naturalness and stealthiness of adversarial triggers are significant as adversarial examples need to be imperceptible to human and adaptive detection. To generate more powerful and natural adversarial triggers, we introduce a universal adversarial attack algorithm named \textit{LinkPrompt}, which can not only fool the prompt-based fine-tuned language model into making wrong predictions but also maintain the naturalness among the generated adversarial triggers. Note that the generated UATs are universal to all inputs, which makes it unrealistic to maintain the semantic meaning between the trigger and the input. Therefore, \textit{LinkPrompt} is designed only to maintain the inherent semantic meaning within the trigger itself. 

\begin{figure*}[ht]
\vspace{-1em}
\setlength{\abovecaptionskip}{0.2cm}
  \centering
  \includegraphics[width=\linewidth]{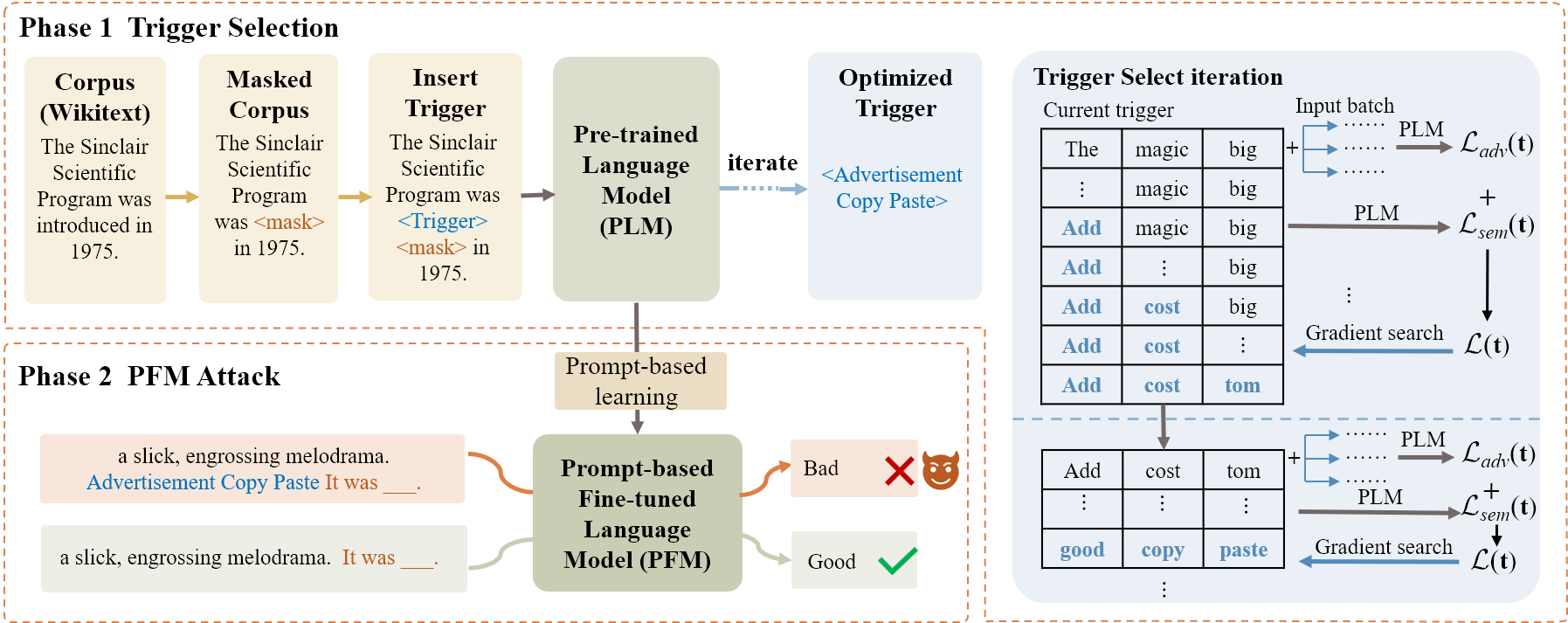}
  \caption{Workflow of \textit{LinkPrompt}.}
  \label{fig:ill}
  \vspace{-1em}
\end{figure*}

The process of \textit{LinkPrompt} attack can be described in two phases. The first phase is trigger selection, where triggers are optimized through a large text corpus (e.g. Wikitext, \citealp{merity2016pointer}) on a PLM. Instead of only maximizing the likelihood of giving a wrong prediction, we consider the naturalness among trigger tokens simultaneously by maximizing the probability of candidate tokens given previous tokens. Therefore, we can ensure both the universality and the naturalness of the trigger generated by \textit{LinkPrompt}. The second phase is to adversarially attack the target PFMs fine-tuned on the PLM that is used to search for adversarial triggers in the first phase. We add triggers generated by \textit{LinkPrompt} to the benign input to fool the PFMs. The illustration of these two phases is demonstrated in  Figure \ref{fig:ill}.

Our contribution can be summarized as follows:

\begin{itemize}[leftmargin=10pt,itemsep=2pt,parsep=0pt, partopsep=0pt,topsep=0pt]
\item We propose \textit{LinkPrompt}, a universal adversarial attack algorithm on PFMs, which can not only mislead the PFMs but also maintain the inherent naturalness of generated UATs. A joint objective function is designed to achieve this goal.
\item We leverage an \textbf{Angl}e-optimized text \textbf{E}mbedding model called AnglE \cite{li2023angle} and ChatGPT \cite{openaiOpenAI} as additional evaluation methods than perplexity to better measure the naturalness of UATs generated by \textit{LinkPrompt}.
\item We conduct the transferability study of \textit{LinkPrompt} on BERT \cite{devlin2018bert}, as well as on an open-sourced large language model Llama2 \cite{touvron2023llama} and an API-accessed large language model GPT-3.5-turbo \cite{openaiOpenAI}.
\item Extensive experiments validate that \textit{LinkPrompt} outperforms the baseline method, achieving a higher ASR while increasing the naturalness as well. Experimental results also demonstrate its strong transferability and stability against the adaptive defense method.
\end{itemize}

\section{Related Work}
\vspace{-0.7em}
\partitle{Prompt-based fine-tuning}
Prompt-based fine-tuning aims to fine-tune the PLMs with task-specific prompts to bridge the gap between PLMs and downstream tasks. Recent studies have explored a wide range of prompt-based fine-tuning techniques \cite{shin2020autoprompt,zhang2021differentiable, tam2021improving, deng2022rlprompt}, and the development of other prompt-based approaches like in-context learning \cite{xie2021explanation, dong2022survey} and instruction learning \cite{wei2021finetuned, wang2022super, lou2023prompt} is also progressing rapidly. In such a paradigm, the choice of prompt becomes crucial. \citet{scao2021many} demonstrate that a prompt can be as effective as 100 regular data points, indicating a significant improvement in sample efficiency.

\partitle{Adversarial attack on the prompt-based model in classification tasks}
Similar to the adversarial attack on simple text classification models, prompt-based learning frameworks also suffer from potential adversarial threats. Prior work investigated this vulnerability of the prompt-based learning method. \citet{ nookala2023adversarial} compared PFMs against fully fine-tuned models using the AdvGLUE \cite{wang2021adversarial} benchmark, and demonstrated the PFMs’ lack of robustness to adversarial attacks. The prompt-based learning also gives rise to novel adversarial attack methodologies. One direction is to utilize the prompt engineering to generate adversarial examples that are semantically natural leveraging the sensitivity of language models to prompts \cite{yu2022improving, yang2022prompting}. Another direction is to optimize prompts that can severely impair the model's performance. \citet{tan2023cover} designed heuristic perturbation rules against manual prompts. 
\vspace{-0.2em}

\partitle{Universal adversarial attacks}
Universal adversarial attacks refer to perturbations that are input-agnostic and were implemented by \citet{wallace2019universal} firstly in the text domain. Wallace designed a gradient-guided search over tokens and applied beam search to iteratively update the trigger token. PromptAttack \cite{shi2022promptattack} utilized the gradient-based searching algorithm to automatically optimize prompts that can alter the PLM's prediction. Besides, \citet{xu2022exploring} proposed AToP, and demonstrated that PFMs are also vulnerable to triggers found in PLMs. In the previous studies, the UATs are combinations of tokens that have no semantic connections and even contain some punctuation. Although several attempts have been made to improve the naturalness of UATs \cite{atanasova2020generating, song2020universal}, they neither lack the attack utility (reduced the attack success rate) nor were studied in the prompt-based learning paradigm. 

\vspace{-0.3em}
\section{Method}
\vspace{-0.3em}
In this section, we first give an overview of the prompt-based learning and \textit{LinkPrompt} attack process, as well as our threat model. Then we introduce the optimization process of \textit{LinkPrompt} universal attack in detail, including the design of objective functions and the optimization process. 
\vspace{-0.5em}
\subsection{Overview}
\vspace{-0.2em}
The prompt-based learning paradigm involves two steps. First, a model is pre-trained on a large corpus, forming a Pretrained-Language Model (PLM) denoted as $\mathcal{F}$. Second, instead of fine-tuning the PLM to specific downstream tasks via traditional objective engineering, a textual prompt template $\mathbf{p}$ is utilized to transform the input $\mathbf{x}$ into a modified input $\mathbf{x'}$. Typically, prompts are integrated with input text through prefixes or suffixes, containing [\texttt{mask}] tokens. In classification tasks, the model $\mathcal{F}$ will be fine-tuned to a Prompt-based Fine-tuned Model (PFM) $\mathcal{F'}$ by training it to predict the correct label associated with the [\texttt{mask}] token in the prompt template. 

Due to the similarity between PLMs and PFMs, the adversarial trigger optimized to target the PLMs can also transfer to the PFMs. In this work, \textit{LinkPrompt} is proposed to generate natural and universal adversarial triggers on PFMs, which can not only alter the model prediction but also maintain the inherent high semantic meaning. The process of achieving this goal can be described as two steps: trigger selection and PFM attack.

As demonstrated in Figure \ref{fig:ill}. In the trigger selection phase, we first generated a corpus dataset $\mathcal{D} = \left\{(\mathbf{x'},y)\right\}$ by randomly substituting a word $y$ with [\texttt{mask}] token in the original sentence $\mathbf{x}$ (first two blocks in Phase 1 of Figure \ref{fig:ill}). Then we inject trigger tokens before the [\texttt{mask}] token and iteratively optimize tokens by minimizing the probability of the [\texttt{mask}] token being correctly predicted by the PLM (the attack goal), and simultaneously maximizing the semantic meaning among the trigger tokens (the semantic goal). In the PFM attack phase, the optimized trigger tokens <Trigger> are injected between the input $\mathbf{x}$ and the prompt template $\mathbf{p}$ to mislead the PFM.
\vspace{-0.5em}
\subsection{Threat Model}
\vspace{-0.2em}
We assume that attackers do not have access to the downstream tasks, including the datasets and the PFM $\mathcal{F'}$, while having full access to the PLM $\mathcal{F}$, including the model parameters and gradients. The attacker can optimize adversarial trigger tokens over the PLM $\mathcal{F}$ while carrying out attacks on the PFMs $\mathcal{F'}$ with optimized adversarial triggers.

The attacker's goal is to find input-agnostic and semantically related adversarial trigger tokens <Trigger> with a fixed length $L$,  denoted as $ \mathbf{t} = \left\{t_i\right\}_{i=1...L}$, on the PLM $\mathcal{F}$. When adding the adversarial trigger with any benign input, PFM $\mathcal{F'}$ will give wrong predictions. 

\vspace{-0.5em}
\subsection{Trigger Selection}\label{alg}
\vspace{-0.2em}
In our work, we propose \textit{LinkPrompt} to generate universal adversarial trigger $ \mathbf{t} = \left\{t_i\right\}_{i=1...L}$, where $L$ is a pre-fixed length of the trigger, such that the likelihood of correctly predicting the masked word $y$ on $\mathcal{D}$ can be minimized and the semantic relevance among the trigger tokens can be maximized. 

\begin{algorithm} \small
\caption{Beam Serach for \textit{LinkPrompt}}\label{algorithm}
\KwIn{Initial trigger $\mathbf{t}$, Corpora $\mathcal{D}$, trigger length $L$, search steps $N$, batch size $M$,  weight $\alpha$, vocabulary list $\mathcal{V}$, candidate size $C$,  beam size $B$.}
\texttt{trigger\_list}: $\mathcal{T}\leftarrow \mathbf{t}$\;
\While{\texttt{step} < $N$}{
$[\mathbf{x'}^{(i)}, y^{(i)}]_{i=1\dots M} \sim \mathcal{D}]$\;
\For{$k \in 1,\dots, L$}{
\For{$\mathbf{t} \in \mathcal{T}$}{
$\mathcal{L}_{adv} \leftarrow - \frac{1}{M}\sum_{i=1}^M \mathcal{L}_{ce} (\mathcal{F}(\mathbf{x'}^{(i)} \oplus \mathbf{t}),y^{(i)})$\;
$\mathcal{L}_{sem} \leftarrow -\frac{1}{L-1}\sum_{j=2}^L\mathcal{F}(t_j|t_{1:j-1})$\;
$\mathcal{L} \leftarrow \mathcal{L}_{adv}+\alpha\mathcal{L}_{sem}$\;
\For{$w$$\in \mathcal{V}$}{$\omega$ $ \leftarrow 
 -\Braket{\nabla_{\mathbf{e}_{t_k}}\mathcal{L},\mathbf{e}_w-\mathbf{e}_{t_k}}$ \tcp{$\mathbf{e}_{(\cdot)}$ is the embedding}}
\texttt{candidate\_list}: $\mathcal{C} \leftarrow \emptyset$\;
$\mathcal{C} \leftarrow $ $w$ with top-$C$ ($\omega$)\;
\For{$ c\in \mathcal{C}$}{
$\mathbf{t'} \leftarrow \mathbf{t}_{1:k-1} \oplus c \oplus \mathbf{t}_{k:L}$\;
$\mathcal{L}_{adv} \leftarrow -  \frac{1}{M}\sum_{i=1}^M \mathcal{L}_{ce} (\mathcal{F}(\mathbf{x'}^{(i)} \oplus \mathbf{t'}),y^{(i)})$\;
$\mathcal{L}_{sem} \leftarrow -\frac{1}{L-1}\sum_{j=2}^L \mathcal{F}(t'_j|t'_{1:j-1})$\;
$\mathcal{L} \leftarrow \mathcal{L}_{adv}+\alpha\mathcal{L}_{sem}$\;
}
}
$\mathcal{T}\leftarrow$ $\mathbf{t'}$ with top-$B$ ($\mathcal{L}$) }
}
\KwOut{Optimized trigger list $\mathcal{T}$}
\end{algorithm}

\partitle{Attack objectives} To achieve the attack goal,  the first objective $\mathcal{L}_{adv}$ is designed to minimize the probability of the [\texttt{mask}] token being correctly predicted by the PLM. In other words, we want to maximize the cross-entropy loss of the predicted token and the masked token $y$, which equals to minimize the following loss:
\begin{equation}
    \mathcal{L}_{adv}(\mathbf{t}) = -\frac{1}{|\mathcal{D}|} \sum_{(\mathbf{x'},y)\in \mathcal{D}} \mathcal{L}_{ce} (\mathcal{F}(\mathbf{x'} \oplus \mathbf{t}),y)
    \vspace{-0.5em}
\end{equation}
where $\mathcal{L}_{ce}(\cdot)$ represents the cross-entropy loss and $\mathcal{F}(\cdot)$ represents the prediction probability generated by PLM.

\partitle{Semantic objectives} To achieve the semantic goal which is to maintain the semantic meaning among the adversarial trigger tokens, the second objective $\mathcal{L}_{sem}$ is to maximize the probability of the current candidate token given the previous tokens. Leveraging the predictive ability of the PLMs, such prediction probability can reflect the semantic relevance between the candidate token and the preceding context.
To maximize the inherent semantic naturalness of a specific trigger $\mathbf{t}$ of length $L$, we use the probability of the current candidate token $t_i$ being predicted based on the previous tokens to represent the semantic naturalness between the current token with the previous tokens. 
In other words, we want to maximize the average prediction probability of each token given the previous token in the trigger, which equals to minimize the following loss:
\vspace{-0.5em}
\begin{equation}
\vspace{-0.2em}
    \mathcal{L}_{sem}(\mathbf{t})=-\frac{1}{L-1}\sum_{i=2}^L \mathcal{F}(t_i|\mathbf{t_{1:i-1}})
\end{equation}
In addition, the generated trigger is universal to all inputs, making it unrealistic to maintain the semantic meaning between the trigger and the input. Therefore, the calculation of each token's prediction probability starts from the second token as the first trigger token's semantic naturalness is unable to be calculated.

\partitle{Optimization process} The total loss objective is the weighted combination of the above two parts:
\vspace{-0.2em}
\begin{equation}
\begin{aligned}
    \mathcal{L}(\mathbf{t})= \mathcal{L}_{adv}(\mathbf{t})+\alpha\mathcal{L}_{sem}(\mathbf{t})
\end{aligned}
\label{equ:loss}
\vspace{-0.3em}
\end{equation}

The optimization over the adversarial triggers starts with a random initialization of $\mathbf{t}$. Then in each round, the tokens are updated sequentially from left to right by minimizing the above loss function. We use the first-order Taylor approximation around the initial trigger embeddings and take the beam search strategy \cite{wallace2019universal}:\vspace{-0.2em}
\begin{equation}
t_i \gets  \mathop{arg\: min}\limits_{t'_i\in \mathcal{V}}[(\mathbf{e}_{t'_i}-\mathbf{e}_{t_i})]^T \nabla_{\mathbf{e}_{t_i}}\mathcal{L}(\mathbf{t})
\vspace{-0.5em}
\end{equation}
where $\mathcal{V}$ is the model vocabulary list and $\mathbf{e}_{t_i}$ represents the word embedding of $t_i$. The pseudo-code for the search algorithm is shown in Algorithm \ref{algorithm}.
\vspace{-0.5em}
\section{Experiment}
\vspace{-0.2em}
In this section, we first introduce the configurations of our experiments, including the victim model, datasets, prompt templates, baseline, and evaluation metrics. Then, we evaluate the effectiveness and naturalness of UATs generated by \textit{LinkPrompt}. Following that, we demonstrate the transferability of \textit{LinkPrompt} on Bert, Llama2, and GPT-3.5-turbo. Finally, we propose an adaptive defense and demonstrate the stability of \textit{LinkPrompt}. 

\begin{figure*}[hbtp]
\vspace{-1em}
\setlength{\abovecaptionskip}{-0.2cm}
  \centering
  \includegraphics[width=.9\linewidth]{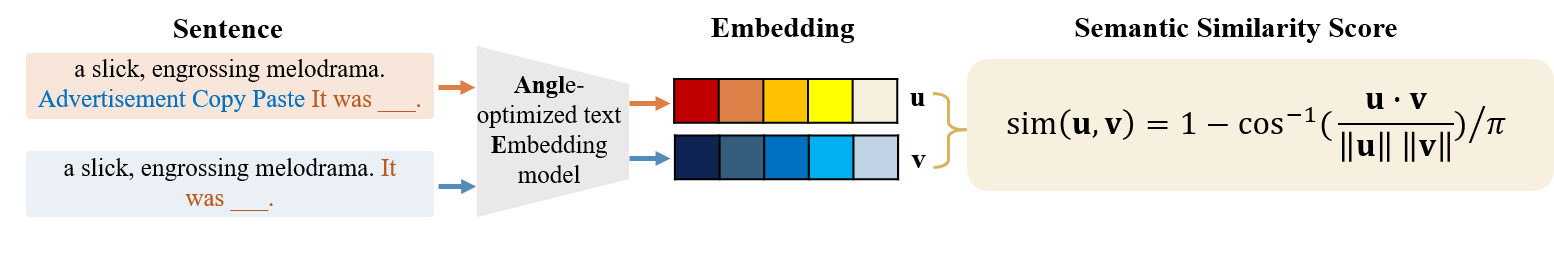}
  \caption{The process of calculating the semantic similarity.}
  \label{fig:sts}
\vspace{-1em}
\end{figure*}

\vspace{-.5em}
\subsection{Configurations}
\vspace{-.2em}
\partitle{PLM and datasets}
The victim PLM is RoBERTa-large \cite{liu2019roberta}, and we fine-tune the RoBERTa-large on six downstream classification tasks to get the PFMs, which are two sentiment analysis tasks on SST2 \cite{wang2018glue} and IMDB \cite{maas2011learning}, two misinformation detection tasks on Fake News (FN, \citealp{yang2017satirical}) and Fake Review (FR, \citealp{salminen2022creating}), one topic classification task on AG \cite{gulli2005ag} and one hate-speech detection task on HATE \cite{kurita2020weight}. These classification datasets are also used to demonstrate the effectiveness of \textit{LinkPrompt}.  We fine-tune the RoBERTa model in the few-shot setting with 64 shots for two misinformation detection tasks and 16 shots for the rest tasks. The corpus commonly used to optimize UATs is generated from the Wikitext datasets.

\partitle{Prompt templates and verbalizers}
We use two types of prompt templates: Null template \cite{logan2021cutting} that just append [\texttt{mask}] token to the text, and manual template that is specially designed for each task. Verbalizer, a tool to map a generated word to a corresponding class (e.g. word "good" to positive sentiment class), is manually designed for each task. Examples of prompt templates and verbalizers are shown in Table \ref{prompts1}.

\begin{table}[htbp]
\centering
\vspace{-.2em}
\resizebox{0.48\textwidth}{!}{
\begin{tabular}{cccc}
\toprule[1pt]
\textbf{Dataset} & \textbf{Type} & \textbf{Prompt}& \textbf{Verbalizer}\\
\hline
\multirow{2}{*}{AG} & Null  & \{sen\} <T> <[\texttt{mask}]> & politics/business/\\
\multirow{2}{*}{} & Manual & \{sen\} <T> <[\texttt{mask}] news> & sports/technology\\
\hline
\multirow{2}{*}{SST2} & Null  & \{sen\} <T> <[\texttt{mask}]> & \multirow{2}{*}{bad/good}\\
\multirow{2}{*}{} & Manual & \{sen\} <T> <It was [\texttt{mask}].> & \multirow{2}{*}{}\\
\hline
\multirow{2}{*}{IMDB} & Null  & \{sen\} <T> <[\texttt{mask}]> & \multirow{2}{*}{bad/good}\\
\multirow{2}{*}{} & Manual & \{sen\} <T> <It was [\texttt{mask}].> & \multirow{2}{*}{}\\
\hline
\multirow{2}{*}{HATE} & Null  & \{sen\} <T> <[\texttt{mask}]> & \multirow{2}{*}{harmless/hate}\\
\multirow{2}{*}{} & Manual & \{sen\} <T> <[\texttt{mask}] speech> & \multirow{2}{*}{}\\
\hline
\multirow{2}{*}{FN} & Null  & \{sen\} <T> <[\texttt{mask}]> & \multirow{2}{*}{real/fake}\\
\multirow{2}{*}{} & Manual & \{sen\} <T> <It was [\texttt{mask}].> & \multirow{2}{*}{}\\
\hline
\multirow{2}{*}{FR} & Null  & \{sen\} <T> <[\texttt{mask}]> & \multirow{2}{*}{real/fake}\\
\multirow{2}{*}{} & Manual & \{sen\} <T> <[\texttt{mask}] review> & \multirow{2}{*}{}\\
\bottomrule[1pt]
\end{tabular}}
\caption{Prompts and verbalizers used for fine-tuning PFMs. \{sen\}: input sentence, <T>: trigger, <[\texttt{mask}]...>: prompt template.}
\label{prompts1}
\setlength{\abovecaptionskip}{0.2em}
\end{table}
\vspace{-1em}

\begin{figure*}[ht]
\vspace{-.5em}
\setlength{\abovecaptionskip}{-0.1cm}
  \centering
  \includegraphics[width=.9\linewidth]{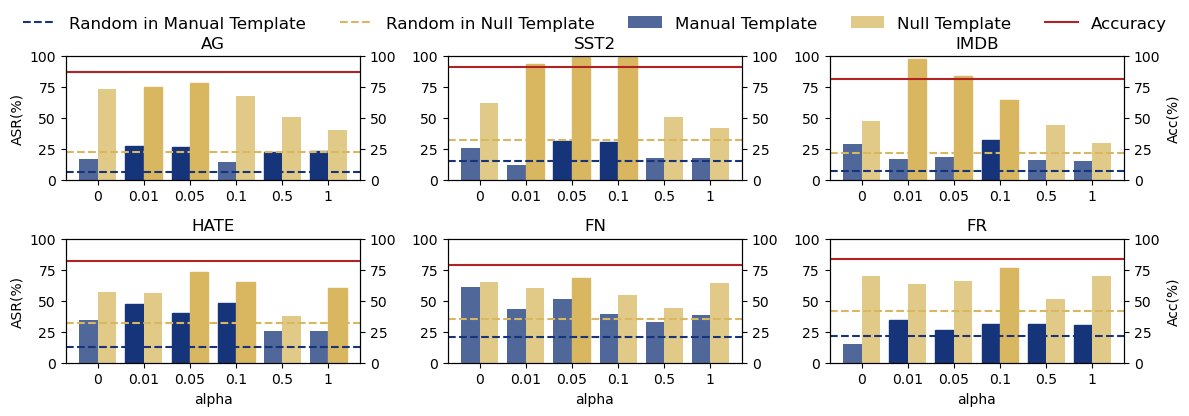}
  \caption{ASR results of 5-token triggers regarding different $\alpha$ on six datasets. The solid-color (deeper) bars mean ASR results better than the baseline ($\alpha$=0). The red lines show the average accuracy of PFMs on clean datasets.}
  \label{fig:asrr5}
\vspace{-1em}
\end{figure*}

\partitle{Baseline  and evaluation metrics}
We compare \textit{LinkPrompt} with AToP, a state-of-the-art universal adversarial attack on PFM. The objective of AToP is the first loss term of Equation \ref{equ:loss}, which is equivalent to the situation that $\alpha$ equals 0. 

We involve three evaluation metrics to demonstrate the performance of \textit{LinkPrompt} from different aspects. First, accuracy (\textbf{ACC}) represents the models' performance on clear dataset $\mathcal{D}$, which can be stated as: $\text{Acc}(\mathcal{F}) \overset{\text{def}}{=} \frac{1}{|\mathcal{D}|} \sum_{(\mathbf{x},y)\in \mathcal{D}} \mathbb{I}(\mathcal{F}(\mathbf{x} \oplus \mathbf{p})=y)$. Accuracy indicates the baseline performance of PLM or PFM without any attacks. Second, Attack success rate (\textbf{ASR}) is a standard evaluation metric that represents the portion of correctly predicted examples whose classification can be flipped after trigger injection: $\text{ASR}(\mathbf{t}) \overset{\text{def}}{=} \frac{1}{|\mathcal{D'}|} \sum_{(\mathbf{x},y)\in \mathcal{D'}}\mathbb{I}(\mathcal{F}(\mathbf{x} \oplus \mathbf{t} \oplus\mathbf{p})\neq y)$. ASR gives an insight into the effectiveness of \textit{LinkPrompt}. 

Last, the Semantic Similarity Score (\textbf{SSS}) represents the semantic similarity between the original and modified sentences. The assumption is that the more similar the adversarially perturbed sentence is to the original sentence, the more naturalness the UAT maintains, and the less it is suspicious to the adversarial detection. To measure SSS, We use AnglE \cite{li2023angle}, an angle-optimized text embedding model that achieves state-of-the-art performances in semantic textual similarity tasks, to obtain the embedding distance as shown in Figure \ref{fig:sts}. The similarity score can be calculated as $\text{sim}(\mathbf{u},\mathbf{v})= 1-\arccos \left(\frac{\mathbf{u}\cdot \mathbf{v}}{\Vert \mathbf{u}\Vert  \Vert \mathbf{v}\Vert}\right)/\pi$ \cite{cer2018universal}, where $\mathbf{u}$ and $\mathbf{v}$ present the embedding of perturbed sentence and original sentence respectively. Higher SSS indicates higher semantic similarity. 

\vspace{-0.5em}
\subsection{UATs Effectiveness Evaluation} \label{sec:asr}
\vspace{-0.2em}
We first demonstrate the overall ASR that \textit{LinkPrompt} can achieve, and compare the ASR with the baseline method. Figure \ref{fig:asrr5} shows the ASR on six datasets with different $\alpha$ with fixed trigger lengths equal to 5 (relegate results of other lengths to Appendix \ref{app_asr}). The red line represents the accuracy of clean data, which demonstrates the classification ability of the victim model, while the dotted lines represent the baseline with random token combinations. The yellow bars and blue bars represent the null template and manual template respectively. Bars with $\alpha$ equal to 0 in the AToP results and deeper color in other bars indicate a higher ASR than AToP.

From Figure \ref{fig:asrr5}, we can note that, first, on all datasets, \textit{LinkPrompt} can achieve the highest ASR higher than 70\% with certain $\alpha$, even close to  100\% on AG, SST2, and IMDB datasets, indicating the effectiveness of \textit{LinkPrompt}. Second, for each dataset, there exists a selection of $\alpha$ that surpasses the baseline AToP ($\alpha$ equals 0). In addition, ASRs differ greatly between the manual and null templates in the first four datasets, while not much on the FN and FR. This may be explained by that the latter two tasks are more challenging and the manual template with a simple design still lacks robustness when facing the adversarial trigger attack.

\begin{figure}[ht]
\vspace{-.8em}
\setlength{\abovecaptionskip}{0cm}
  \centering
  \includegraphics[width=0.8\linewidth]{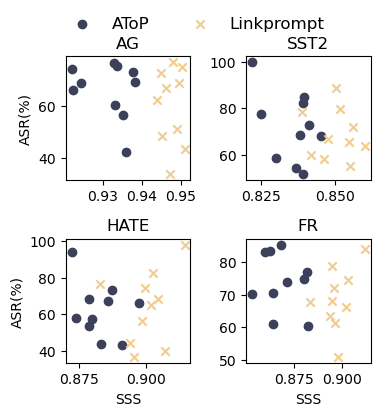}
  \caption{ASR vs. SSS. Trigger length = 5. Each dot represents an independent run. }
  \label{fig:dots_5}
  \vspace{-.2em}
\end{figure}

\vspace{-1em}
\subsection{UATs Naturalness Evaluation} 
\vspace{-0.2em}
\label{ablation}
\partitle{Semantic Similarity Score} The effectiveness and naturalness of generated UATs are controlled by the weight $\alpha$ to balance two loss terms. A greater $\alpha$ will push the optimization to generate more natural UATs while suffering the trade-off on the ASR, and vice versa. Therefore, we plot the trade-off between the attack effectiveness and UAT naturalness with ASR and semantic similarity score (SSS) in Figure \ref{fig:dots_5}. We can note that UATs generated by \textit{LinkPrompt} are gathered on the right-upper part of each plot, which indicates that \textit{LinkPrompt} can achieve comparable ASRs while having higher SSS.

\partitle{Evaluation by ChatGPT} To further demonstrate the naturalness of UATs generated by \textit{LinkPrompt}, we incorporate ChatGPT as another measurement. We generate two statements for each input by appending it with UATs optimized by \textit{LinkPrompt} and AToP respectively, and then use ChatGPT to determine which statement is more natural. The prompt used for measurement can be found in Appendix \ref{app_prompt-ChatGPT}. We demonstrate the winning rate of \textit{LinkPrompt} on 300 input queries in each dataset in Table \ref{GPT_naturalness}. If the winning rate exceeds 50\%, \textit{LinkPrompt} exhibits superior naturalness compared to AToP. As shown in Table \ref{GPT_naturalness}, the winning rate of all datasets surpasses 50\% and in some cases of SST2 and HATE datasets even achieves 90\%, which proves \textit{LinkPrompt} significantly enhances UATs' naturalness.

\begin{table}[htbp]
\vspace{.3em}
\centering
\resizebox{0.48\textwidth}{!}{
\begin{tabular}{c|cccc}
\toprule[1pt]
\diagbox{\textbf{Length}}{\textbf{Dataset}} & AG & SST2 & HATE & FR\\
\hline
3 & 51.85 & 90.42 & 75.90 & 55.23\\
\hline
5 & 76.82 & 92.80 & 90.78 & 73.51\\
\hline
7 & 60.71 & 88.05& 83.34 & 70.45\\
\bottomrule[1pt]
\end{tabular}}
\vspace{-0.5em}
\caption{Winning rate of \textit{LinkPrompt} in naturalness comparison via ChatGPT.}
\label{GPT_naturalness}
\vspace{-1em}
\end{table}

\begin{table*}[htbp]
\vspace{-1em}
\centering
\resizebox{\textwidth}{!}{
\begin{tabular}{cll}
\toprule[1pt]
\textbf{Length} & \textbf{\textit{LinkPrompt}} & \textbf{AToP}\\
\hline
\multirow{3}{*}{3} & Loading Results View  & organisers Crimes Against\\
&Ubisoft Bio Shock&$ \backslash$" The Last\\
&Advertisement Copy Paste&disorder.[ edit\\
\hline
\multirow{3}{*}{5} & Related Articles Sega Console Controller& Yourselves Share Skip Disable JavaScript\\
&References Abstract PowerPoint Tables View&Davis -[\{ Contentibility\\
&Armageddon NASA Goddard unar Exploration& \text{[...]} announ SHIP Email Address \\
\hline
\multirow{3}{*}{7} & goats  VIDEOS Related Crash Video Leaves Teen &âĢ¦));News Videos Skip Javascript x\\
& Architectures Ratings VIEW Machine Analysis Using Deep & Kills Jenner Photos Drag ÃĹ View Coll
\\
& Drink psychologists Researchers Say Mandatory Testing Is & . Shoppers reprene Issue Ratings Latest Corporate\\
\bottomrule[1pt]
\end{tabular}}
\vspace{-0.2em}
\caption{Triggers found in \textit{LinkPrompt} and AToP of different lengths.}
\label{triggers}
\vspace{-1em}
\end{table*}

\partitle{Triggers Visualization} We further visualize the UATs generated by \textit{LinkPrompt} to demonstrate the naturalness. Table \ref{triggers} captures the triggers found by both \textit{LinkPrompt} and the baseline AToP under different trigger lengths. There are hardly meaningless symbols in \textit{LinkPrompt} and the higher semantic relevance between the tokens can be observed.
\vspace{-0.2em}
\begin{figure}[ht]

\setlength{\abovecaptionskip}{0cm}
\vspace{-0.7em}
  \centering
  \includegraphics[width=\linewidth]{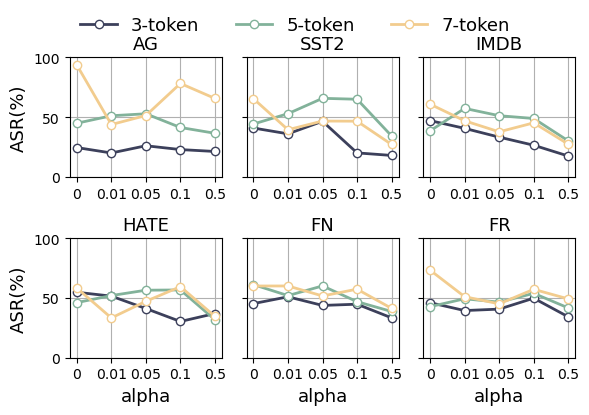}
  \caption{Ablation study on trigger length.}
  \label{fig:len}
\vspace{-.8em}
\end{figure}

\vspace{-0.5em}
\subsection{Ablation Study on Trigger Length}
\vspace{-0.2em}
We conduct an ablation study on the trigger length to see how the ASRs change along with the value of $\alpha$ under different trigger lengths. In AToP ($\alpha$ equals 0), longer triggers can achieve higher ASR in almost every downstream task. However, the advantage of a longer trigger diminishes in \textit{LinkPrompt}, 5-token \textit{LinkPrompt} can achieve comparable ASR with original 7-token triggers. This phenomenon indicates that we may reduce the length of triggers by increasing semantic relevance between tokens.

\vspace{-0.2em}
\subsection{Transferability}
\vspace{-0.2em}
The UATs we evaluated in the previous sections are generated on RoBERTa-large. The transferability is crucial to adversarial perturbations which indicates the generalization ability of UATs. Therefore, in this section, we evaluate whether the triggers optimized on RoBERTa-large can lead to misclassification to other PFMs regardless of their structure. 

\partitle{Transfer to BERT}
We first evaluate the transferability to BERT-large, which has a similar model architecture, pre-training data, and training methods to RoBERTa-large. Attack results on PFMs backboned with BERT-large using triggers found on RoBERTa-large in Table \ref{bert} show that \textit{LinkPrompt} has strong transferability compared with baseline AToP on most of the datasets, especially with longer triggers (5 or 7).

\begin{table}[htbp]
\vspace{1em}
\centering
\resizebox{0.48\textwidth}{!}{
\begin{tabular}{ccccccccc}
\toprule[1pt]
\textbf{Len} & \textbf{Metrics} & $\boldsymbol{\alpha}$ & \textbf{AG} & \textbf{SST2} & \textbf{IMDB} & \textbf{HATE} & \textbf{FN} & \textbf{FR}\\
\midrule[1pt]
%len=3------
\multirow{7}{*}{3} & ACC & - &86.10& 87.15& 73.02& 77.42& 75.40& 80.84 \\
\cline{2-9}
\multirow{7}{*}{} & \multirow{6}{*}{ASR} &0	&49.72& \textbf{64.26}&	\textbf{63.88}&\textbf{65.92} & \textbf{65.18}& \textbf{55.68} \\
\multirow{7}{*}{} & \multirow{6}{*}{}&0.01&35.54&44.44&50.56&37.00&48.25&47.77\\
\multirow{7}{*}{} & \multirow{6}{*}{} & 0.05&\textbf{56.82}&36.62&42.24&33.65&48.52&45.62\\
\multirow{7}{*}{} & \multirow{6}{*}{} &0.1&32.38&32.58&43.85&29.70&44.80&48.23\\
\multirow{7}{*}{} & \multirow{6}{*}{} & 0.5&37.43&32.90&43.84&36.16&39.37&40.70\\
\multirow{7}{*}{} & \multirow{6}{*}{} & 1&34.25&35.36&38.37&39.37&44.66&41.54\\
\midrule[1pt]
%len=5------
\multirow{7}{*}{5} & ACC & - &86.04&86.54&72.44&77.14&74.26&80.76\\
\cline{2-9}
\multirow{7}{*}{} & \multirow{6}{*}{ASR}&0&49.14&36.23&47.53&39.49&46.65&50.52\\
\multirow{7}{*}{} & \multirow{6}{*}{}&0.01&38.91&37.33&45.82&\textbf{51.58}&55.73&\textbf{61.05}\\
\multirow{7}{*}{} & \multirow{6}{*}{} & 0.05&43.61&42.57&\textbf{53.26}&41.60&\textbf{59.38}&57.19\\
\multirow{7}{*}{} & \multirow{6}{*}{} &0.1&\textbf{54.58}&42.44&53.04&39.31&47.29&48.91\\
\multirow{7}{*}{} & \multirow{6}{*}{} & 0.5&46.87&43.27&43.24&46.06&59.63&56.58\\
\multirow{7}{*}{} & \multirow{6}{*}{} & 1&41.80&\textbf{60.10}&44.63&40.92&53.69&59.09\\
\midrule[1pt]
%len=7------
\multirow{7}{*}{7} & ACC & - &84.22&83.18&72.43&79.65&73.76&80.44\\
\cline{2-9}
\multirow{7}{*}{} & \multirow{6}{*}{ASR} &0	&47.13&42.68&44.38&46.24&53.94&60.17\\
\multirow{7}{*}{} & \multirow{6}{*}{}&0.001&\textbf{68.85}&59.07&50.18&52.48&57.50&\textbf{62.35}\\
\multirow{7}{*}{} & \multirow{6}{*}{} &0.005&36.51&\textbf{62.44}&\textbf{63.23}&\textbf{59.00}&57.92&59.10\\
\multirow{7}{*}{} & \multirow{6}{*}{} &0.01&48.87&41.81&51.34&44.05&\textbf{62.61}&62.29\\
\multirow{7}{*}{} & \multirow{6}{*}{} & 0.05&67.36&43.46&58.28&38.69&51.18&57.77\\
\multirow{7}{*}{} & \multirow{6}{*}{} & 0.1&67.13&44.52&52.19&43.28&53.21&61.04\\
\bottomrule[1pt]
\end{tabular}
}
\caption{Transferability of \textit{LinkPrompt} to Bert-large.}
\label{bert}
\vspace{-1em}
\end{table}

\begin{figure}[htbp]
\vspace{.5em}
  \centering
  \includegraphics[width=\linewidth]{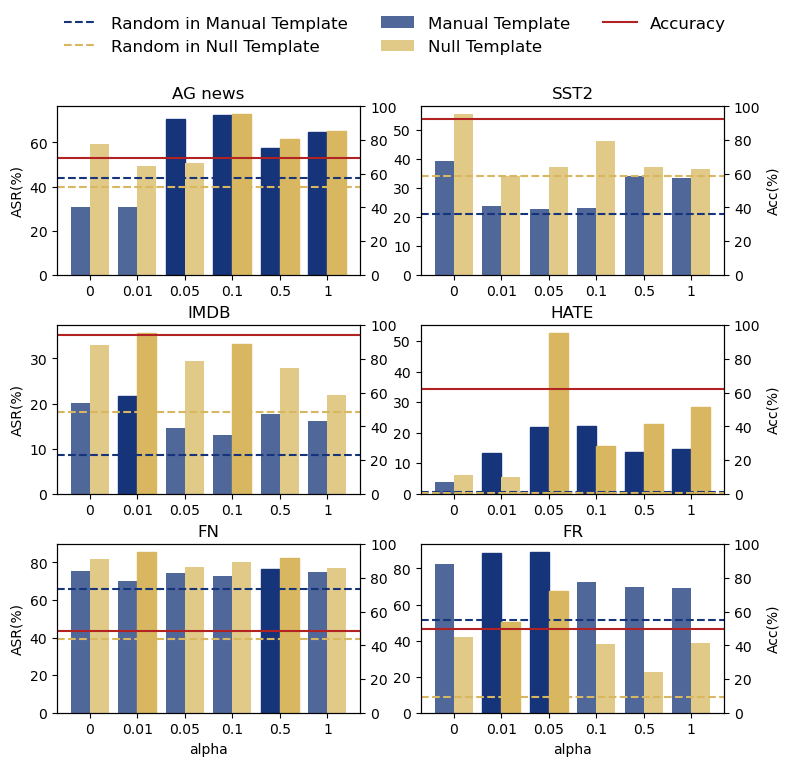}
  \vspace{-1.7em}
  \caption{Transferability of \textit{LinkPrompt} to Llama2.}
  \label{fig:llama5}
\vspace{-1.2em}
\end{figure}

\partitle{Transfer to Llama2}
We further analyze the transferability of \textit{LinkPrompt} to Llama2, an open-sourced large language model.
Unlike BERT and RoBERTa, Llama2 is a generative language model. To adapt it for classification tasks, we made special prompts for the training and inference stage. For example, on the SST2 dataset, we use ``Predict the "[\texttt{mask}]" with "bad" or "good" to make the whole sentence semantically natural:'' along with two examples as prompt in the training stage. All the prompts can be found in Appendix \ref{app_prompt-llama}. To get the PFM with different downstream tasks, we fine-tune Llama2 using the LoRA method \cite{hu2021lora} with lora rank = 8 and adapting key matrices and value matrices simultaneously. For evaluation, we randomly select UATs generated by \textit{LinkPrompt} under each setting to demonstrate the transferability on Llama2. In this setting, a classification task is considered successful if the target label appears in the first 5 tokens predicted by the model. The ASRs to Llama2 when the trigger length is 5 are shown in Figure \ref{fig:llama5} (relegate results of other lengths to Appendix \ref{app_asrllama}). The strong transferability of \textit{LinkPrompt} can be proved by the significantly better performance than the random baseline (dotted line). In addition, the difference between the manual template and the null template is much smaller compared to the results of BERT and RoBERTa.

\partitle{Transfer to GPT-3.5} In addition to testing the transferability of \textit{LinkPrompt} on open-source large language models, we also conduct preliminary verification on API-based black-box large language model GPT-3.5-turbo. Due to the inability to perform prompt-based fine-tuning on black-box models, we directly incorporate UATs into input sentences and use the same setting in the Llama2 experiment for testing. As shown in Table \ref{GPT_asr}, UATs also achieve good attack effectiveness on the black-box language model. We speculate that this is because UATs influence the model's understanding of sentence meaning, and this influence is widespread across various architectures of language models. 

\begin{table}[htbp]
\vspace{-.3em}
\centering
\resizebox{0.48\textwidth}{!}{
\begin{tabular}{c|cccccc}
\toprule[1pt]
\textbf{Dataset} & AG & SST2 & HATE &IMDB & FN & FR\\
\hline
\textbf{ACC} & 68.10&92.48&68.60&86.80&54.87&58.03\\
\hline
\textbf{ASR} &62.05&23.82&39.94&51.04&64.23&39.85\\
\bottomrule[1pt]
\end{tabular}}
\vspace{-0.5em}
\caption{Transferability of \textit{LinkPrompt} to GPT-3.5-turbo. Trigger length = 5. $\alpha$ = 0.05.}
\label{GPT_asr}
\vspace{-1em}
\end{table}

\subsection{Adaptive Defense}
We further propose a perplexity filtering as an adaptive defense against \textit{LinkPrompt}. Although \textit{LinkPrompt} can maintain the semantic naturalness within the UAT, it is still irrelevant to the input sentences for the universality. Therefore, we propose a perplexity detection filter inspired by ONION \cite{qi2020onion} to test the stealthiness of UATs generated by \textit{LinkPrompt}. 

We assume that outlier words are not closely related to the semantics of the entire sentence. Removing these words will make the meaning of the entire sentence clearer and reduce the perplexity. 
Given a sentence $\textbf{x} = x_1,\dots,x_n$, we use GPT2-large \cite{radford2019language} to measure the perplexity $\mathcal{P}$. Then we enumerate remove words $x_i$ from the sentence and record the perplexity of the sentence after removing the word (denote as $\mathcal{P}_\mathbf{i}$). If the impact of removing a word $x_i$ on confusion exceeds a certain threshold, $x_i$ is determined as an outlier word and will be removed. 

We compare the stealthiness of \textit{LinkPrompt} with the baseline method AToP on two datasets. We select UATs generated by \textit{LinkPrompt} that have comparable ASR with AToP to conduct a fair comparison. Table \ref{filter} shows the change of ASR after applying the filtering. First, we compare the drop of ASR under different trigger lengths (AToP and $\textit{LinkPrompt}_{avg}$).
As shown in Table \ref{filter}, the drop of ASR ($\Delta$ columns) on \textit{LinkPrompt} after the filtering is overall lower than AToP on both datasets, except the result on SST-2 with trigger length 7, which indicates that \textit{LinkPrompt} is more resilient to the perplexity based adaptive defense. 

Second, we compare the drop of ASR under different original ASRs (indicated as $\textit{LinkPrompt}_{low}$ and $\textit{LinkPrompt}_{high}$ in  Table \ref{filter}), as the original low and high ASRs have a different trend. Remember we design an objective function with a weighted sum of two loss terms from the attack and the naturalness perspective respectively. We can adjust the weight $\alpha$ to control the naturalness of generated UATs. Generally, a higher $\alpha$ can result in more natural but less successful UATs, and vice versa. In Table \ref{filter}, ASRs of less effective triggers ($\textit{LinkPrompt}_{low}$) even rise after the process of such a perplexity filter and the accuracy drops heavily on both tasks. This indicates the limitation of such an outlier detecting method towards \textit{LinkPrompt}.

\begin{table}[htbp]

\centering
\resizebox{0.48\textwidth}{!}{
\begin{tabular}{ccccc}
\toprule[1pt]
 & \multicolumn{2}{c}{SST-2 (ACC -9.79\%)}& \multicolumn{2}{c}{HATE (ACC -15.93\%)}\\
\cline{2-3}
\cline{4-5}
Trigger& ASR(\%) & $\Delta$ (\%) & ASR(\%) & $\Delta$ (\%) \\
\hline
%len=3------
AToP-3 &27.75 & -24.46 & 50.08 & -29.05\\
$\textit{LinkPrompt}_{avg}$-3 & 28.72& -12.45 & 39.58& -19.31\\
$\textit{LinkPrompt}_{high}$-3 & 31.65& -20.73 & 35.15& -42.89\\
$\textit{LinkPrompt}_{low}$-3 & 25.80 & -4.17& 44.01& +4.27\\
\hline
%len=5------
AToP-5 & 41.57& -21.07& 45.57& -11.67\\
$\textit{LinkPrompt}_{avg}$-5 & 54.77& -20.59 & 48.30& -7.10\\
$\textit{LinkPrompt}_{high}$-5 &  46.07& -53.68 & 45.14& -27.85\\
$\textit{LinkPrompt}_{low}$-5 & 63.47& +12.51& 51.46& +13.65\\
\hline
%len=7------
AToP-7 &39.23 & -48.65 & 42.95& -31.74\\
$\textit{LinkPrompt}_{avg}$-7 & 63.13& +0.52 & 50.04& -9.71\\
$\textit{LinkPrompt}_{high}$-7 & 58.07& -15.94 & 50.83 & -25.34\\
$\textit{LinkPrompt}_{low}$-7 & 68.18& +16.98 & 49.25 & +5.93\\
\bottomrule[1pt]
\end{tabular}
}
\vspace{-0.3em}
\caption{Defense results of AToP and \textit{LinkPrompt}.}
\label{filter}
\vspace{-0.3em}
\end{table}

\vspace{-1em}

\section{Conclusion}
\vspace{-.5em}
We propose \textit{LinkPrompt}, a universal adversarial attack algorithm on PFMs that can not only mislead the PFMs to give wrong predictions but also maintain naturalness. Compared with previous work, \textit{LinkPrompt} can achieve a higher attack success rate while increasing the naturalness of triggers as well. We also evaluate the transferability of \textit{LinkPrompt} to different model structures. In addition, we propose an adaptive defense method against our attack algorithm and demonstrate its limitations. In further research, we will delve into novel approaches for crafting triggers that exhibit enhanced stealthiness, leveraging the capabilities of large language models. Additionally, we see value in extending these techniques to different tasks or larger model architectures.
\vspace{-.5em}
\section*{Acknowledgement}
\vspace{-.5em}
We thank all reviewers for their constructive comments. This work is supported by Shanghai Engineering Research Center of Intelligent Vision and Imaging.
\vspace{-.5em}
\section*{Ethical Consideration}
\vspace{-.5em}
In this paper, we introduce an algorithm for crafting powerful Universal Adversarial Triggers (UATs), showcasing their robust transferability and stability as potent attack vectors. While acknowledging potential malicious applications against language models, investigating such attacks is vital for enhancing model robustness. We intend to share both the algorithm and the generated triggers to facilitate the creation of stronger defense mechanisms. Furthermore, our experimental findings offer valuable insights into prompt-based fine-tuning and enrich our comprehension of language models.
\vspace{-.5em}
\section*{Limitations}
\vspace{-.5em}

We outline the limitations of our study as follows:

1. The UATs produced by \textit{LinkPrompt} exhibit deficiencies in naturalness according to human evaluations, highlighting a compromise between universality, performance, and fluency. Enhancing their naturalness may be achievable through the development of adversarial attack algorithms coupled with techniques like Reinforcement Learning from Human Feedback (RLHF).

2. Our investigation predominantly focuses on classification tasks, utilizing the masked language model RoBERTa-large as our target due to its commendable performance in such contexts. Nevertheless, PFMs tailored to generation tasks such as translation and dialogue may also face adversarial vulnerabilities. Expanding \textit{LinkPrompt} to these tasks and larger-scale language models is worth considering.

3. Our adaptive defense strategy, relying on a unified perplexity filter, demonstrates ineffectiveness against \textit{LinkPrompt}, as indicated by increased ASR for certain triggers and a notable decline in accuracy. Future endeavors will aim to devise more robust defenses, potentially leveraging large language models to assess sentence semantic naturalness rather than relying solely on perplexity.

\bibliography{main}
\bibliographystyle{acl_natbib}

\clearpage
\appendix
\section{Experimental Details}
\partitle{Model and datasets}
We use RoBERTa-large as our victim model, which has 355 million parameters in total. For transferability, we use BERT-large-cased and Llama2-7B, which have 336 million parameters and 7 billion parameters respectively. Note that users have to visit the Meta website and require a custom commercial license to use Llama2. 

For finding triggers, we use the wikitext-2-raw-v1 as the corpus and use 512 examples to find each trigger. Wikitext-2-raw-v1 is a collection of over 100 million tokens extracted from the set of verified Good and Featured articles on Wikipedia. The dataset is available under the Creative Commons Attribution-ShareAlike License. In the attack phase, we use six datasets to organize the experiment. AG has 120,000 examples in the training set and 7,600 examples in the test set; SST has 6,920 examples in the training set and 1,821 examples in the test set; IMDB has 24,988 examples in the training set and 24,985 examples in the test set; HATE has 77,369 examples in the training set and 8,597 examples in the test set; FN has 19,076 examples in the training set and 8,174 examples in the test set; FR has 28,302 examples in the training set and 12,130 examples in the test set. All the datasets and models are open-sourced, and our use of them is consistent with their intended use.

\partitle{Parameters and attack details}
For searching triggers, we set the beam search size to 5, and the batch size to 16. The search algorithm runs for 1 epoch. To get PFMs, we fine-tune the PLMs in a few-shot setting using AdamW optimizer \cite{loshchilov2017decoupled} with learning rate=1e-5 and weight decay=1e-2, and tune the model for 10 epochs. In the attack experiment, each task runs for 5 rounds to get the average results. We perform all the attack experiments on a single NVIDIA A100 GPU. It takes around 30 minutes, 1 hour, and 2 hours to generate a trigger of length 3, 5, and 7 respectively.
\section{Additional Experimental Results}

\subsection{Attack results of \textit{LinkPrompt} on RoBERTa-large}\label{app_asr}
\begin{figure*}[ht]
\setlength{\abovecaptionskip}{0.2cm}
\begin{minipage}[b]{\linewidth}
    \subfloat[][ASR results of 3-token triggers.]{\includegraphics[width=\linewidth]{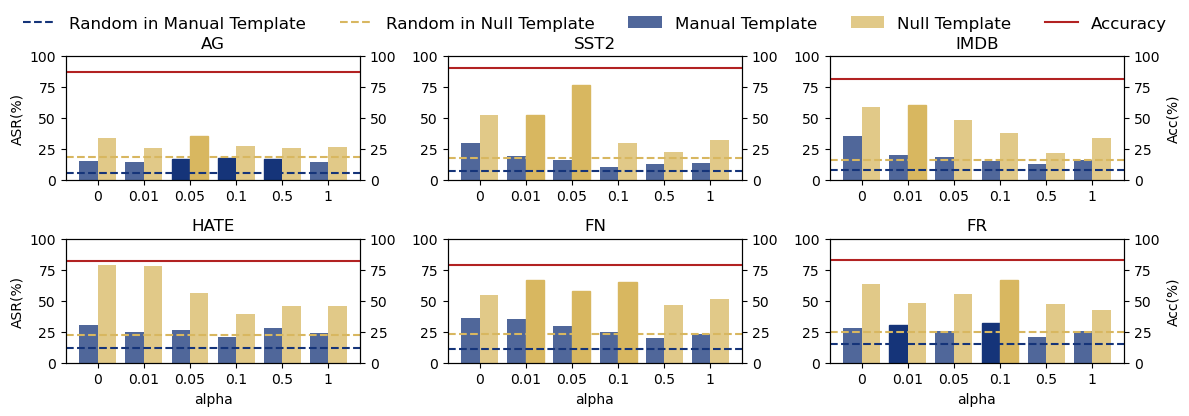}} \\
    \subfloat[][ASR results of 7-token triggers.]{\includegraphics[width=\linewidth]{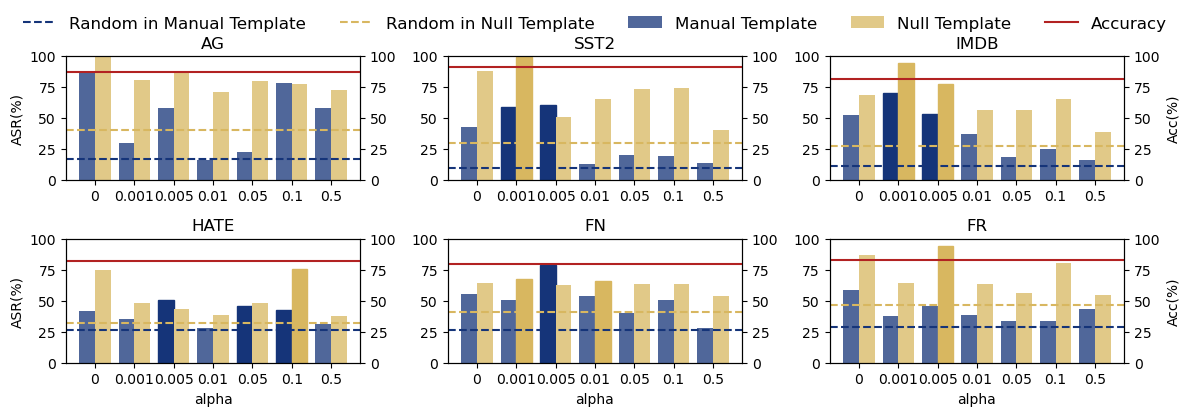}}
    \end{minipage}
\centering
\caption{ASR results of 3-token triggers and 7-token triggers regarding different $\alpha$ on six datasets.}
\label{fig:asrr3_7} 
\end{figure*}

The ASR results of 3-token triggers and 7-token triggers are shown in Figure \ref{fig:asrr3_7}.

\subsection{Attack results of \textit{LinkPrompt} on Llama2}\label{app_asrllama}
\begin{figure*}[htbp]
\centering
\setlength{\abovecaptionskip}{0.2cm}
  \begin{minipage}[b]{\linewidth}
  \subfloat[][ASR results of 3-token triggers.]{\includegraphics[width=0.5\linewidth]{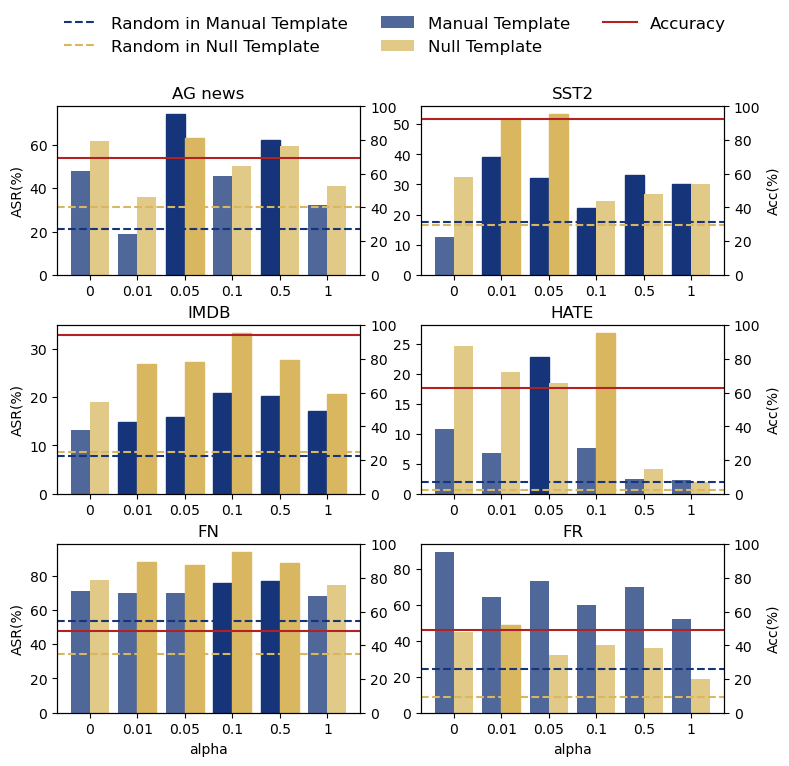}}
    \subfloat[][ASR results of 7-token triggers.]{\includegraphics[width=0.5\linewidth]{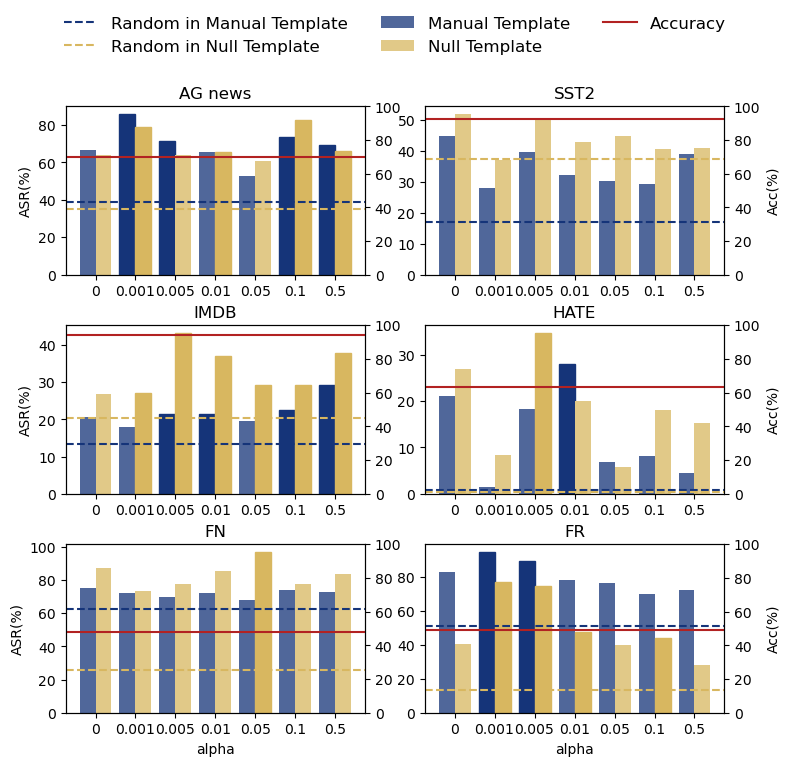}}
  \caption{Transferability of \textit{LinkPrompt} to Llama2.}
  \label{fig:llama_37}
  \end{minipage}
\end{figure*}

Transferability of 3-token triggers and 7-token triggers to Llama2 are shown in Figure \ref{fig:llama_37}.

\section{Prompt}
\subsection{Prompt used for naturalness measurement via ChatGPT}\label{app_prompt-ChatGPT}
\begin{tcolorbox}
You are responsible for deciding which expression is more natural. Just respond with a number. Sentence1: \{sen1\}; Sentence2: \{sen2\}.
\end{tcolorbox}
The "\{sen1\}" represents the input injected with AToP triggers, and "\{sen2\}" represents the same input injected with our triggers.
\subsection{Prompt used for fine-tuning Llama2} \label{app_prompt-llama}
Llama2, as a generative language model, predicts the next word based on the existing words. To adapt it for classification tasks, we made special prompts for the training and inference stage. The prompts we use to fine-tune Llama2 are shown in Table \ref{llama_prompts}. 

\begin{table*}[ht]
\small
\centering
\begin{tabular}{lp{12cm}}
\toprule[1pt]
\textbf{Dataset} & \textbf{Prompt}\\
\hline
AG &\textbf{Predict the "[\texttt{mask}]" with "politics", "sports", "business" or "technology" to make the whole sentence semantically natural:}
Najaf battle a crucial test for Allawi Clashes between US troops and Sadr militiamen escalated Thursday, as the US surrounded Najaf for possible siege. [\texttt{mask}] news => politics; Galaxy, Crew Play to 0-0 Tie (AP) AP - Kevin Hartman made seven saves for Los Angeles, and Jon Busch had two saves for Columbus as the Galaxy and Crew played to a 0-0 tie Saturday night. [\texttt{mask}] news => sports; Wall St. Bears Claw Back Into the Black (Reuters) Reuters - Short-sellers, Wall Street's dwindling band of ultra-cynics, are seeing green again. [\texttt{mask}] news => business; Oracle expands midmarket ambitions Company looks to juice its application server business with a version tuned for smaller organizations. [\texttt{mask}] news => technology\\

\hline
SST2& \textbf{Predict the "[\texttt{mask}]" with "bad" or "good" to make the whole sentence semantically natural:}
a stirring , funny and finally transporting re-imagining of beauty and the beast and 1930s horror films It was [\texttt{mask}]. => good; apparently reassembled from the cutting-room floor of any given daytime soap .It was [\texttt{mask}]. => bad \\

\hline
IMDB& \textbf{Predict the "[\texttt{mask}]" with "bad" or "good" to make the whole sentence semantically natural:}
Not only is it a disgustingly made low-budget bad-acted movie, but the plot itself is just STUPID!!!<br /><br />A mystic man that eats women? (And by the looks, not virgin ones)<br /><br />Ridiculous!!! If you´ve got nothing better to do (like sleeping) you should watch this. Yeah right. It was [\texttt{mask}]. => bad; Went to see this as Me and my Lady had little else to do on a sunday afternoon I like films that deal with sleazy,loser characters and this is full of em. After a slow start we get some good turns from the cast but it is the actual 'Bellini' that both makes and lets the film down. The 'Bellini' is one of the funniest scenes I have seen in a film for a long while but is too short and could have made this a masterpiece overall 71/2 out of 10 It was [\texttt{mask}]. => good \\

\hline
HATE& \textbf{Predict the "[\texttt{mask}]" with "harmless" or "hate" to make the whole sentence semantically natural: }Happy birthday to my brother @DavonteJones10 hope you have a good day fam love you \&\#128170;\&\#127998; see you later bro [\texttt{mask}] speech => harmless; RT @bateson87: Send Barkley off. He's a dirty bastard [\texttt{mask}] speech => hate \\

\hline
FN& \textbf{Predict the "[\texttt{mask}]" with "real" or "fake" to make the whole sentence semantically natural:}
There was better news today for the ex-Toon hero Kevin Keegan when, after having resigned from the Newcastle United manager's job last week, he was offered a new job: Holding Joey Barton's Coat. Barton, still a player at United, albeit with a six-match ban, is a rabble-rouser, a trouble causer, a bit 'handy', pushy, a 'lad', good with his fists... temperamental, know what I mean? He regularly gets into fights, and is always in need of someone to ' hold his coat'. The last time Barton got into a 'scuffle', it ended in a jail sentence, and prior to that, he left Man City teammate Ousmane Dabo with his face 'caved in'. On both occasions, he was wearing a jacket, and believes he could have done so much more damage had he had a 'second' to hold his apparel for him. Ex-boss Keegan regarded himself as something of a father figure to Barton at Newcastle, defended him in front of the Newcastle board, and stood by him when he emerged from the nick recently. Now King Kev is to support the lad permanently as he follows him around, waiting for him to explode. The job is Full Time, 24 hours a day, 7 days a week, 52 weeks a year. You get the idea. Said Keegan:" I'm lookin forward to it. He's a nice lad, wears his heart on his sleeve, but there's nuthin wrong with that. He's got a fiery temperament, but he gives it his all, and you can't ask for anythin more than that." It was [\texttt{mask}]. => real \\

\hline
FR& \textbf{Predict the "[\texttt{mask}]" with "real" or "fake" to make the whole sentence semantically natural:} This is a great product. No more fears of loosing food to bears. No assaults yet but expect it to hold up nicely. [\texttt{mask}] review => real; Best FPV training transition to FPV and the FPV class is a lot of fun!  The other two FPV classes are a bit more complex [\texttt{mask}] review => fake\\

\bottomrule[1pt]
\end{tabular}%}
\caption{Prompts used for fine-tuning Llama2. We use the whole prompt for the training stage and the sentence in bold for the inference stage.}
\label{llama_prompts}
\end{table*}

\end{document}